\pdfoutput=1

\documentclass[11pt]{article}

\usepackage[]{emnlp2022}

\usepackage{times}
\usepackage{latexsym}
\usepackage{todonotes}
\usepackage[T1]{fontenc}

\usepackage{graphicx}
\usepackage{afterpage}

\usepackage{subfig}

\usepackage[utf8]{inputenc}

\usepackage{microtype}

\title{Discourse Relation Embeddings: \\Representing the Relations between Discourse Segments in Social Media}

\author{Youngseo Son \hspace{1.25em} Vasudha Varadarajan \hspace{1.25em} H. Andrew Schwartz \\
  Department of Computer Science, Stony Brook University \\
  \texttt{\{yson,vvaradarajan,has\}@cs.stonybrook.edu}}

\begin{document}
\maketitle
\begin{abstract}

Discourse relations are typically modeled as a discrete class that characterizes the relation between segments of text (e.g. causal explanations, expansions).  
However, such predefined discrete classes limit the universe of potential relations and their nuanced differences.
Adding higher-level semantic structure to modern contextual word embeddings, we propose representing discourse relations as points in high dimensional continuous space. 
However, unlike words, discourse relations often have no surface form (relations are \textit{inbetween} \textit{two segments}, often with no explicit word or phrase marker), presenting a challenge for existing embedding techniques. 
We present a novel method for automatically creating \textit{discourse relation embeddings} (DiscRE), addressing the embedding challenge through a weakly supervised, multitask approach. 
Results show DiscRE representations obtain the best performance on Twitter discourse relation classification (macro $F1=0.76$) and social media causality prediction (from $F1=.79$ to $.81$), performing beyond modern sentence and word transformers, and  capturing novel nuanced relations (e.g. relations at the intersection of causal explanations and counterfactuals). 

\end{abstract}

\section{Introduction}
\begin{figure}[t!]
\centering
  \includegraphics[width=3.0in]{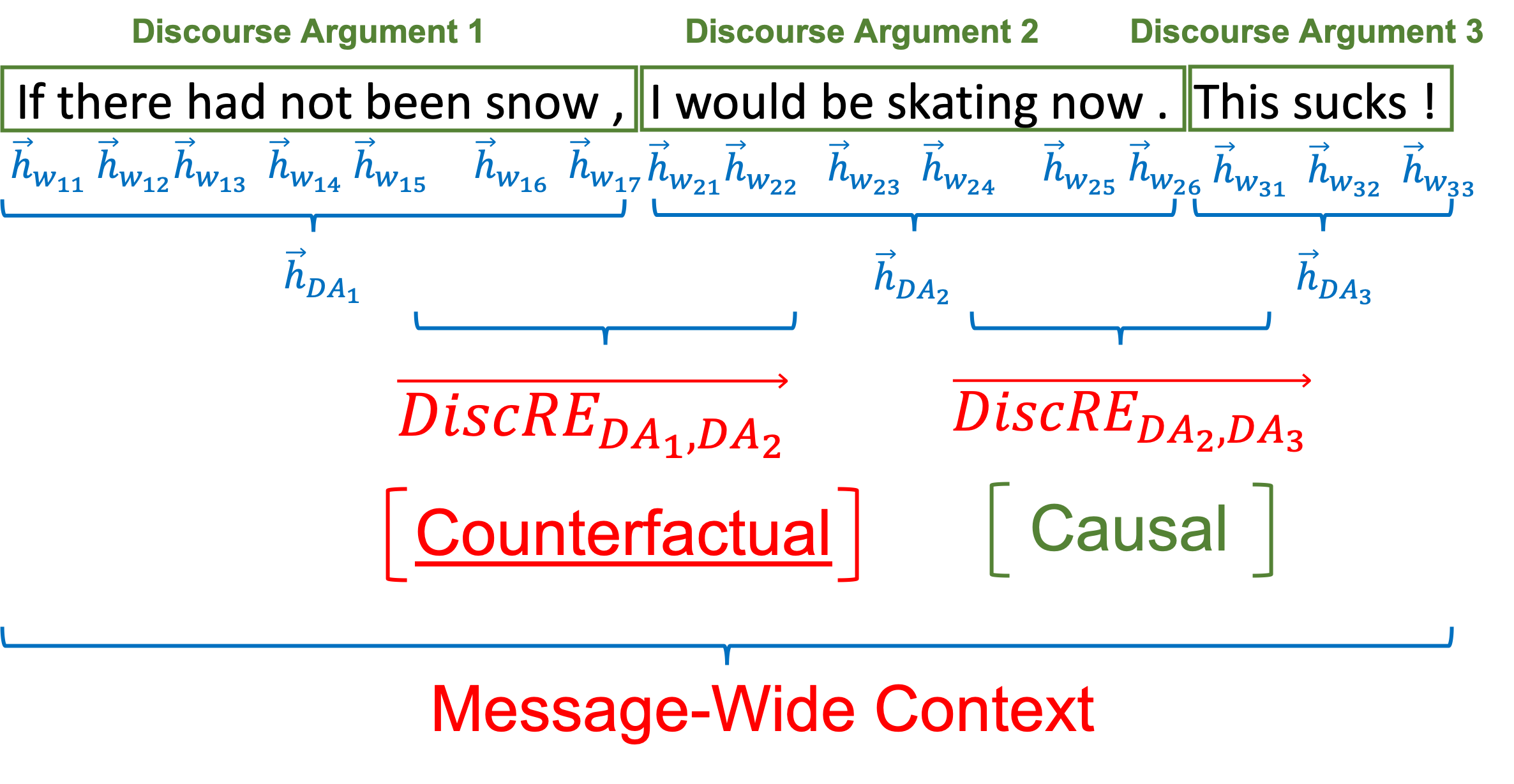}
  \caption{Our model DiscRE predicts relations of adjacent discourse arguments based on other text spans of the whole message as context. By learning and embedding fine-grained properties of discourse relation with the posteriors from PDTB into a continuous vector space, DiscRE may learn existing discourse relation tagsets like `causal' relations, but also new latent discourse relations such as `counterfactual' relations.}
  \label{fig:parsing_comapre}
\end{figure}

Relations between discourse segments (i.e., phrases rooted by a main verb phrases or clauses) have mostly been studied as discrete classes; most notably Penn Discourse Treebank (PDTB)~\cite{prasad2008penn} and Rhetorical Structure Theory Discourse Treebank (RST DT)~\cite{carlson2001building} contain 43 and 72 types of discourse relations respectively.
At the same time, such work has taken place over newswire, the domain of both the PDTB and RST.
With many different relation classes over sophisticated schema, annotation is non-trivial prohibiting extensive development in new domains (e.g., social media). 
Thus, progress in developing, training and evaluating discourse relation identifiers has happened over \textit{discrete-class} models with \textit{labeled} \textit{newswire} corpora~\cite{pitler2009automatic,park2012improving,ji2014representation,lin2014pdtb,popa2019implicit}.

To address this challenge and enable expansion of discourse work to social media, we propose a weakly supervised learning method which does not require any explicit labels.
Instead, it adds a semantic structure that can effectively capture various types of discourse relations, even in other domains leveraging a multitask learning method called ``Discourse Relation Embeddings (DiscRE)''. Our DiscRE model represents discourse relations as continuous vectors rather than single discrete classes. 

As the first study of \textit{embedding} discourse relations into high dimensional continuous spaces, we mainly focus on social media. Social media is a challenging domain because it contains many acronyms, emojis, unicode, and informal variations of grammatical structure, but its personal nature provides diverse and psychologically-relevant discourse patterns which are not often found from newswire text. 
According to our best knowledge, there are only relatively small datasets for specific types of discourse relations for causal relation~\cite{son2018causal} and counterfactual relations~\cite{son2017recognizing}, but they are not diverse and large enough to learn general discourse relations.


In this paper, we propose a novel weakly supervised learning method for deriving discourse relation embeddings on social media. We created a social media discourse relation dataset and validated our new approach. Furthermore, we conducted visual investigations on continuous discourse relation spaces and thorough qualitative analysis on the behaviors of DiscRE in both PDTB and social media. Then, we also validated how well our learning method can generalize across different domains by applying DiscRE as transfer learning features for discourse relation downstream tasks. 

Our contributions include: (1) a novel model structure which, when weakly supervised, creates embeddings capturing discourse relations (DiscRE), (2) the creation of new Twitter discourse relation dataset and the validation of our approach for the discourse relation classification on the dataset, (3) quantitative and qualitative evaluation of DiscRE on PTDB and downstream social media discourse relation tasks in which DiscRE outperformed strong modern contextual word and sentence embeddings, obtaining a new state-of-the-art performance for causality and counterfactuals, and (4) the release of all of our datasets and models.

\section{Related Work}
Our work builds on previous studies in discourse relations with two key distinctions: (1) the predominant set of work on discourse relations has focused on annotated newswire datasets (PDTB and RST DT) rather than social media; (2) work to improve discourse parsing has focused either on feature engineering or models for better predicting \textit{predefined} discourse relations rather than embeddings (or latent relations).
Such work takes pre-segmented clauses as input~\cite{pitler2009automatic,park2012improving} or builds full end-to-end discourse parsers~\cite{ji2014representation,lin2014pdtb}.  
\citet{kishimoto-etal-2020-adapting} looked into adapting BERT for relation classification by pretraining with domain text and connective prediction. 
Other methods have zeroed-in on implicit discourse relations (those without a connective token) and also used a hierarchical model but for discourse classification rather than embedding \citep{Bai2018DeepER}. Some work from \citet{varia-etal-2019-discourse, ma-yan-2021, zhang-etal-2021-context} leverage CNNs and graph networks to capture relationships between adjacent discourse units for implicit discourse relation classification.

Some have studied \textit{single} discourse relations over social media. 
\newcite{son2017recognizing} used a hybrid rule-based and feature based supervised classifier to capture counterfactual statements from tweets. 
\newcite{bhatia2015better} and \newcite{ji2017neural} applied RST discourse parsing to social media movie review sentiment analysis, showing a pretrained model which was optimized for RST DT, suffered from domain differences when it was run on different domains (e.g., legislative bill). 
\citet{son2018causal} developed a causal relation extraction model using hierarchical RNNs to parse social media. In general, hierarchical RNN-based models have worked well in general for capturing specific relations in social media and other discourse relations outside social media~\cite{bhatia2015better,son2018causal,ji2017neural}.

Our work is related to modern multi-purpose contextual word embeddings~\cite{devlin2018bert,peters2018deep} in the motivation to utilize latent representations in order to capture context-specific meaning. However, our model generates contextual discourse relation embeddings by learning probabilities rather than discrete labels and, it can learn all possible relations even from the same text leveraging posterior probabilities from well-established study~\cite{prasad2008penn}.

We also build on research that has assembled custom discourse relation datasets or created training instances from existing datasets using discourse connectives~\cite{jernite2017discourse,nie2019dissent,sileo2019mining}. \citet{jernite2017discourse} designed an objective function to learn discourse relation categories (conjunction) based on discourse connectives along with other discourse coherence measurements while \citet{nie2019dissent} and \citet{sileo2019mining} used objectives to predict discourse connectives. Here, we devised an objective function for learning posterior probabilities of discourse relations of the given discourse connectives, so the model can capture more fine-grained senses and discourse relation properties of the connectives\footnote{e.g., `since' can signal a temporal relation in `I have been working for this company since I graduated', but might signal a causal relation `I like him since he is very kind to me'.}. Also, all of them used sentence encoders to learn sentence representations and compared their learned representations with other state-of-the-art sentence embeddings such as Infersent~\cite{conneau2017supervised}. 
However, our DiscRE model learns a ``discourse relation'' representation (i.e. embedding) between discourse arguments rather than the representation of a respective text span of the pair (Figure~\ref{fig:parsing_comapre}).

Finally, some have studied an RNN-attention-based approach to multitask learning for discourse relation predictions in PDTB~\cite{lan2017multi,ji2016latent} and a sentence encoder with multi-purpose learning for discourse-based objectives~\cite{jernite2017discourse}. Also, \citet{liu2016implicit} leveraged a multi-task neural network for discourse parsing across existing discourse trees and discourse connectives. \citet{shi-demberg-2019-next} used next sentence prediction to get better at implicit discourse relation classification.

A particular challenge of these prior works has been to improve performance when no connective is explicitly mentioned in the text. 
All of these works utilized predefined discrete classes of possible discourse relations. 
While we were inspired and build on some of their techniques, our task is more broadly defined as producing vector representations of the relationship between discourse segments \textit{not limited to predefined discourse relations} (whether defined with explicit connectives or conventional discourse signals exist or not) and is evaluated over a broad diversity of discourse relation tasks as well as downstream applications.

\begin{figure}
    \centering
    \includegraphics[width=3in]{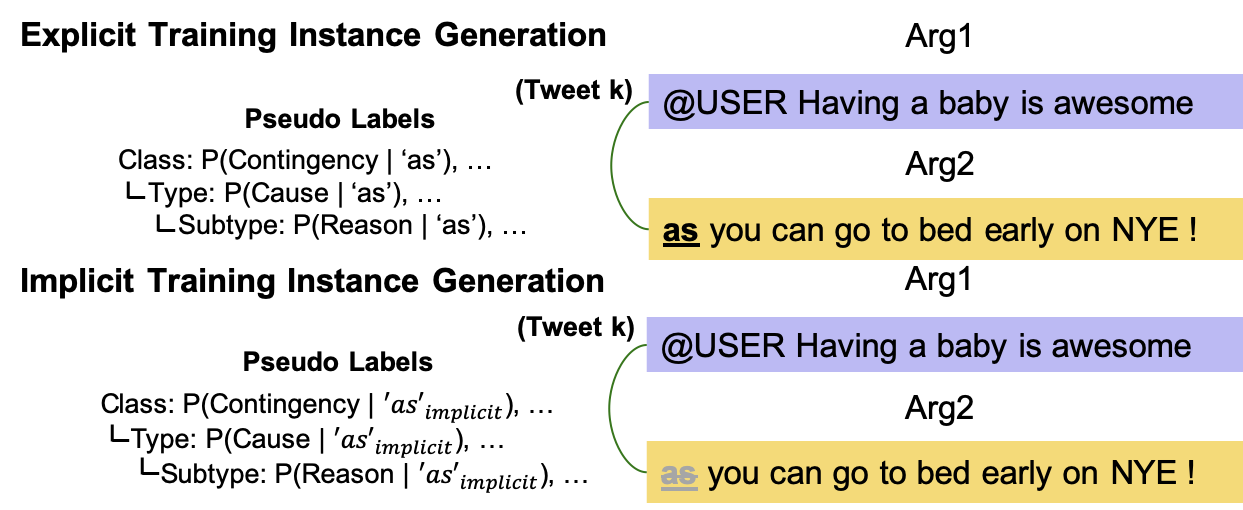}
    \caption{Training instance generation example. For explicit relation training, the training instance is labeled with the posterior probabilities of all possible \textit{Class}, \textit{Type}, and \textit{Subtype} given the explicit connective `as' from PDTB. 
    }\label{fig:training_instance_generation}
\end{figure}

\begin{figure}
    \centering
    \includegraphics[width=3.4in]{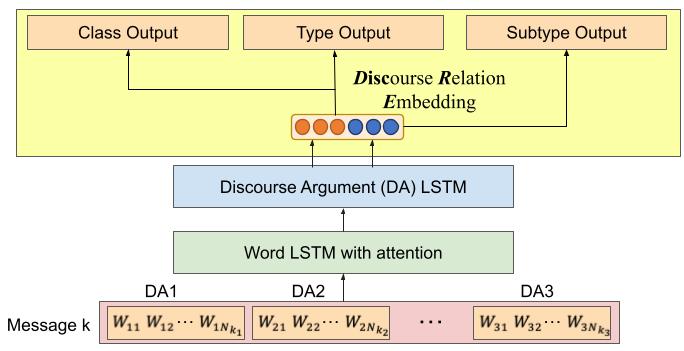}
    \caption{Our model learns different nuances and high dimensional contextual discourse relations by learning probabilities of all possible discourse relations in the relation hierarchy (\textit{Class}, \textit{Type}, and \textit{Subtype}).}
    \label{fig:model_structure}
\end{figure}

\section{Methods}
The base for our model is a hierarchical BiRNN, following work on capturing causal relations in social media~\cite{son2018causal}, but we have added word-level attention, reflecting the necessity to keep word-level markers while parsing higher-order discourse relations (e.g., word pairs, modality, or N-grams)~\cite{pitler2009automatic}. 

\subsection{Data Collection}
\paragraph{DiscRE Weakly-Supervised Learning Training Set.} No existing annotated discourse relation dataset exists for social media. Thus, we collected random tweets from December 2018 through January 2019 for training. Non-English tweets were filtered out, and URLs and user mentions were replaced with separate special tokens respectively. 
For training, we collected only messages which contained at least one of the most frequent discourse connectives from each PDTB discourse sense (\textit{Type}) annotation\footnote{after, before, when, but, though, nevertheless, however, because, if, and, for example, or, except, also.} among random tweets from January 2019: up to 3,000 messages for each type of discourse relation which is similar to the numbers in existing social media discourse relation datasets. 
With this process, we 1) balance our training set to have similar effect sizes of target datasets, 2) minimize potential biases towards a few dominant discourse relations in Twitter, and 3) keep the minimal numbers of discourse relation data samples to validate the effectiveness of the computationally efficient objective function for directly capturing discourse relations. Originally we found 20,787 tweets with our keyword search, but our discourse connective disambiguation process (see details in Section~\ref{sec:da_extarction}) left us 11,517 tweets. We chose random 10\% of them as our development set to tune hyperparameters.

\paragraph{Qualitative Analysis Evaluation Set.}\label{sec:qual_set} For our qualitative analysis, we separately collected 10,000 random tweets from December 2018 without any restrictions so we can test our model on an unseen and unbiased natural social media test set as possible. This setting also allows us to conduct qualitative analysis with minimized potential biases which might exaggerate the capabilities of our model (e.g., our model would be evaluated on discourse relations and discourse connectives it had never seen during its training, so it would not be able to depend only on posterior probabilities of certain discourse connectives used as keywords for training set collection to obtain coherent qualitative analysis results).

\paragraph{PDTB-style Twitter Discourse Relation Dataset.} As an additional social media evaluation, we created a Twitter discourse relation classification dataset. We collected 360 tweets from September 2020 using the same preprocessing methods for DiscRE training set. Specifically, first, we collected 30 tweets using all discourse connectives of each discourse relation class (i.e., \textit{Contingency}, \textit{Temporal}, \textit{Comparison}, and \textit{Expansion}) as search keywords from random tweets, so 120 tweets in total. Then, three well-trained annotators annotated whether each set of 30 tweets have its target relations as a binary classification. 
Finally, we randomly shuffled 120 keyword tweets and 240 non-keyword random tweets, and annotators classified four discourse relation classes. 
Pairwise inter-rater agreement was 85\%, with three-way reliable in the moderate range (Fleiss $\kappa=0.49$).
We used majority vote as our discourse relation labels. Among 360 tweets, there were 36 \textit{Contingency}, 8 \textit{Temporal}, 22 \textit{Comparison}, and 43 \textit{Expansion} relations. The rest of the tweets were annotated with \textit{None}.
\subsection{Discourse Argument Extraction}\label{sec:da_extarction}
We adopted the PDTB-style argument extraction method as it is relatively simple and thus more robust in noisy texts of social media. For argument extraction, we combined approaches of~\citet{biran2015pdtb} and~\citet{son2018causal}. 

We extract the sentences and use the Tweebo parser \citet{kong-etal-2014-dependency} 
to extract discourse arguments (we identified discourse connectives only if there are verb phrases\footnote{minimal discourse units defined in \citet{prasad2008penn}}). If there is discourse connective in a sentence, we identify an argument to which a discourse connective attached as \textit{Arg2} , and the other as \textit{Arg1} \cite{prasad2007penn}. For discourse connectives at the beginning of a tweet, we identify the text from the beginning until the end of the first verb phrase separated by punctuation Tweet POS tags or other discourse connectives as \textit{Arg2}, and the rest as \textit{Arg1}; if a discourse connective or coordinating conjunction Tweet POS tag is in the middle, we identify the text from start to the middle connective as \textit{Arg1}, and from the connective to the end as \textit{Arg2}~\cite{biran2015pdtb}. We also identify emojis as separate discourse arguments as suggested by \cite{son2018causal} since they play a critical role in signaling implicit relations. 


\subsection{Training}
We use weakly supervised multitask learning with a hierarchy of PDTB-style discourse relation learners (Figure~\ref{fig:training_instance_generation}). 
Note that this method, as opposed to entirely self-supervised (i.e. predict next discourse argument), enables us to capture the relationships beyond the likelihood of one discourse argument to appear after another (i.e. how BERT models sentences), which would not necessarily distinguish one relationship from another. 

\paragraph{Pseudo Labeling and Training Instance Generation.}

For each discourse argument pair, the discourse connectives were extracted, and the pair was labelled with all of the possible relations that are found in PDTB. We use the ratio of these possible discourse relations given the discourse connective as a weight within binary cross-entropy loss -- this idea of using probabilistic labels follows the work in \textit{pseudo labeling} for image recognition~\cite{lee2013pseudo}. More specifically, two types of training instances were used for the weakly supervised learning of DiscRE: explicit relation pairs and implicit relation pairs. For explicit relation training pairs, the discourse argument which contains discourse connectives is defined as \textit{Arg2} and the rest text span of the pair is defined as \textit{Arg1}. This segmentation method obtained state-of-the-art performances for previous discourse relation tasks~\cite{biran2015pdtb,son2018causal}. For implicit relation training pairs, the discourse connective is removed from \textit{Arg2} of each pair; \citet{rutherford2015improving} found this approach can learn strong additional signals quite well, although it is not perfectly equivalent to learning implicit discourse relations\footnote{Among the discourse connectives we used for our training, only `if' belongs to the \textit{`Non-omissible'} discourse connective class and even this class showed relatively high effectiveness for implicit relation training when omitted~\cite{rutherford2015improving}.}. 
Next, each of these generated pairs were input along with its whole tweet as its context to our DiscRE model optimize the model towards the objective function to learn the posterior distributions of all possible relations given the discourse connective in PDTB (Figure~\ref{fig:model_structure}). Importantly, this mode of labeling is self scalable, yet it also enables a relatively delicate learning objective which considers all possible discourse relations rather than predicting just discourse connectives. 

\subsection{Discourse Relation Embeddings}

We used a hierarchical bidirectional LSTM model; the first layer LSTM (Word LSTM) captures interaction between words of each discourse argument with attention. The second layer LSTM (Discourse Argument LSTM) captures relations among all discourse arguments across the whole tweet. This architecture was inspired by \citet{son2018causal} and \citet{ji2017neural} as they found that their similar hierarchical model architecture performed well in related discourse relation tasks. As the first work to attempt embedding relations, we choose RNNs because the sequences of discourse units are of a similar size as where RNNs have been successful over transformers elsewhere \cite{matero-schwartz-2020-autoregressive}. Discourse relations, by their definition, describe relations between neighboring or close discourse units, and thus do not have the same motivations for attention-based architectures as long distance dependencies in sequences of words.

This model was optimized on each tweet for training towards the following objective function:
$$J(\theta)=-\sum_{i}\sum_{j=1}^{N_i} w_{ij} y_{ij} log(f_{i}(x_{ij}))) $$
where $i$ is three levels of discourse relation hierarchy from PDTB (\textit{Class}, \textit{Type}, and \textit{Subtype}) and $N_i$ is the dimension of all existing relations in each level and $w_{ij}$ is the posterior from PDTB of the relations given the discourse connective in the current pair of arguments. This can be viewed as multitask learning of shared RNN layers for three different level outputs (Figure~\ref{fig:model_structure}). The hidden vectors of \textit{Arg1} and \textit{Arg2} from Discourse Argument LSTM were concatenated to learn \textit{Class} output and \textit{Type} output, as these are relations between two arguments. Whereas, only the hidden vector of \textit{Arg2} from Discourse Argument LSTM was used for learning \textit{Subtype} as it is rather a role of \textit{Arg2}, given the \textit{Class} and \textit{Type} relations (Figure~\ref{fig:model_structure}). There is a dropout layer with a dropout rate of 0.3 (as suggested in \newcite{ji2017neural} and \newcite{son2018causal}) between Word LSTM and Discourse Argument LSTM.

Finally, for generating DiscRE, the hidden vectors of \textit{Arg1} and \textit{Arg2}, and the output vectors of \textit{Class}, \textit{Type}, and \textit{Subtype} were concatenated. With this structure, DiscRE can capture latent features of discourse relations between any given argument pair, based on the context across all other discourse arguments in addition to probabilities of predefined discourse relations with contextual nuances (Figure~\ref{fig:model_structure}). 

\paragraph{Model Configuration.}
DiscRE is implemented in PyTorch~\cite{paszke2019pytorch}. For hyperparameter tuning, we explored the dimensions of pretrained word embeddings (Glove) 
and hidden vectors 25, 50, 100, and 200 with SGD and Adam~\cite{kingma2014adam}. 
We chose the models which obtain best performances on our development set, which used Adam with 200 dimensions and typically 50 epochs.
We implemented a word-level attention as defined in \cite{yang2016hierarchical} but with ReLU function for its activation. 
We compare with other similar models such as: (1) BERT, for which we used BERT base uncased model (12 layers, 768 hidden dimensions, and 12 heads) by HuggingFace \footnote{\url{https://huggingface.co/bert-base-uncased}}
and (2) InferSent, for which we used a pretrained model trained with 300 dimension glove vectors as inputs and 2,048 LSTM hidden dimensions.

\section{Results}
DiscRE was validated on both newswire and social media discourse relation tasks. Additionally, qualitative analysis on the DiscRE representations were explore for both the domains. 

\subsection{Evaluations}
 First, we examined whether DiscRE can capture discourse relations in PDTB, even though grammatical properties and general text formats of newswire and social media are quite different. Then, we evaluated our model for social media discourse relation tasks: causal relation prediction and Twitter discourse relation classification. We used linear SVMs for all transfer learner classifiers for evaluation as this model obtained the best performance from the previous related work~\cite{son2018causal}.
 
 \begin{table*}[tbh!]
\centering
\begin{tabular}{|l|llll|ll|}
\hline Models & CON. & TEM. & COM. & EXP.  & Mic. & Mac.  \\ \hline 
Ngrams & 0.575 & 0.693 & 0.757 & 0.757 & 0.709 & 0.695 \\
BERT & \textbf{0.612} & 0.724 & 0.746 & 0.748 & 0.714 & 0.708 \\ 
Inferse. & 0.604 & 0.670 & 0.738 & 0.726 & 0.693 & 0.685 \\\hline
\textbf{DiscRE} & 0.598 & \textbf{0.736} & \textbf{0.768} & \textbf{0.768} & \textbf{0.726} & \textbf{0.718}\\

\hline
\end{tabular}
\caption{\label{tab:PDTB_rel} F1 scores of the four-way PDTB discourse class prediction (`CON.': \textit{Contingency}, `TEM.': \textit{Temporal}, `COM.': \textit{Comparison}, `EXP.': \textit{Expansion}). We report both micro F1 and macro F1. DiscRE obtained the best performances across all four discourse relation classes except for the second best performance for Contingency class prediction F1.}
\end{table*}

 \paragraph{Transfer Learning on PDTB.}
In order to measure how well our model can generalize to different domains and capture predefined newswire discourse relations, we conducted transfer learning experiments for predicting the four senses of Level-1 discourse relation classes: \textit{Contingency}, \textit{Temporal}, \textit{Comparison}, and \textit{Expansion}. 

 We  extract DiscRE from the pairs: \textit{Arg1} and \textit{Arg2} from the PDTB dataset, and used them as transfer learning features to a linear classifier. The PDTB dataset was created with the annotators first segmenting the texts into discourse arguments, and then annotating a discourse relation between each pair of neighboring discourse arguments (marked as \textit{Arg1} and \textit{Arg2}). To make a fair comparison, we extracted BERT, Ngrams, and Infersent from \textit{Arg1} and \textit{Arg2} and the concatenation of \textit{Arg1} and \textit{Arg2} to use as separate features, so that the transfer learned model can recognize the notion of \textit{Arg1} and \textit{Arg2} and utilize the whole context as well. The classifiers were trained with each of these embeddings and we report the performances.

As suggested in \newcite{prasad2007penn}, we used Sections 2 to 21 for training and Section 23 for testing in PDTB. Despite the relatively small number of the training set and larger domain differences with newswire target domains in its pretraining procedures, DiscRE still obtained the best performance for overall discourse relation predictions except for \textit{Contingency} classification F1. This may indicate that DiscRE learns domain-agnostic signals for discourse relations leveraging discourse connectives in the weakly supervised multitask learning settings. (Table~\ref{tab:PDTB_rel}).

\paragraph{Causal Relation Prediction on Social Media.} \label{subsec:causal}
We evaluated our model on a causality prediction task on social media messages collected by \newcite{son2018causal}. The DiscRE embeddings of the messages were extracted and for each message, the embeddings were averaged over for the transfer learning features for causality prediction. For comparison, BERT embeddings were also extracted for each discourse unit, and averaged for each message in the dataset, and Infersent sentence embeddings were directly extracted from the messages. The transfer learned classifier from DiscRE embeddings can be used to improve over the best results reported in the previous work on causality prediction~\cite{son2018causal}. DiscRE obtained better performances ($F1=0.752$) than BERT ($F1=0.746$) and Infersent (F1=0.709) and overall, this simple transfer learning approach using obtained a comparable performance to the models used in \newcite{son2018causal} ($F1=0.791$) (Table~\ref{tab:causality}). On further exploration, we found that fine-tuning BERT for the causality prediction task improved the performance to $F1=0.789$. Furthermore, when DiscRE was used along with best performing text features from \newcite{son2018causal} (N-grams, Tweet POS tags, Word Pairs~\cite{pitler2009automatic}, sentiment tags) of the messages for transfer learning, we obtained a new state-of-the-art performance (See Table \ref{tab:causality})

\begin{table}
\centering
\begin{tabular}{|l|l|l|l|}
\hline  Model & F1  \\ \hline
\cite{son2018causal} & 0.791 \\\hline
BERT & 0.746 \\
Infersent  & 0.709  \\
DiscRE & 0.752  \\ \hline
BERT Fine-Tuned & 0.789 \\
\textbf{DiscRE + ALL} & \textbf{0.807} \\

\hline
\end{tabular}
\caption{\label{tab:causality} Causality prediction performance of DiscRE compared to other models. DiscRE-based classifier obtained the new state-of-the-art performance.}
\end{table}

\begin{table*}[tbh!]
\centering
\begin{tabular}{|l|lllll|ll|}
\hline Models & CON. & TEM. & COM. & EXP. & None  & Mic. & Mac.  \\ \hline 
Ngrams & 0.386  & 0.386  & 0.353 & 0.119 & 0.813 & 0.686 & 0.407  \\
BERT & 0.412 & 0.000 & 0.426 & 0.086 & 0.857 & 0.706 & 0.316 \\ 
Inferse. & 0.390  & 0.111 & 0.566 & 0.324 & 0.867 & 0.719 & 0.452 \\

\hline
\textbf{DiscRE} & \textbf{0.478} & \textbf{0.421} & \textbf{0.591} & \textbf{0.400} & \textbf{0.883} & \textbf{0.758} & \textbf{0.554}\\

\hline
\end{tabular}
\caption{\label{tab:Twitter_rel} F1 scores of the discourse class prediction on Twitter (`CON.': \textit{Contingency}, `TEM.': \textit{Temporal}, `COM.': \textit{Comparison}, `EXP.': \textit{Expansion}). Then, we report both micro F1 and macro F1. DiscRE obtained the best performance across all relations.}
\end{table*}

\begin{figure*}[tbh]
\centering
\subfloat[Average Attention Weights on Twitter]{{\includegraphics[width=3in]{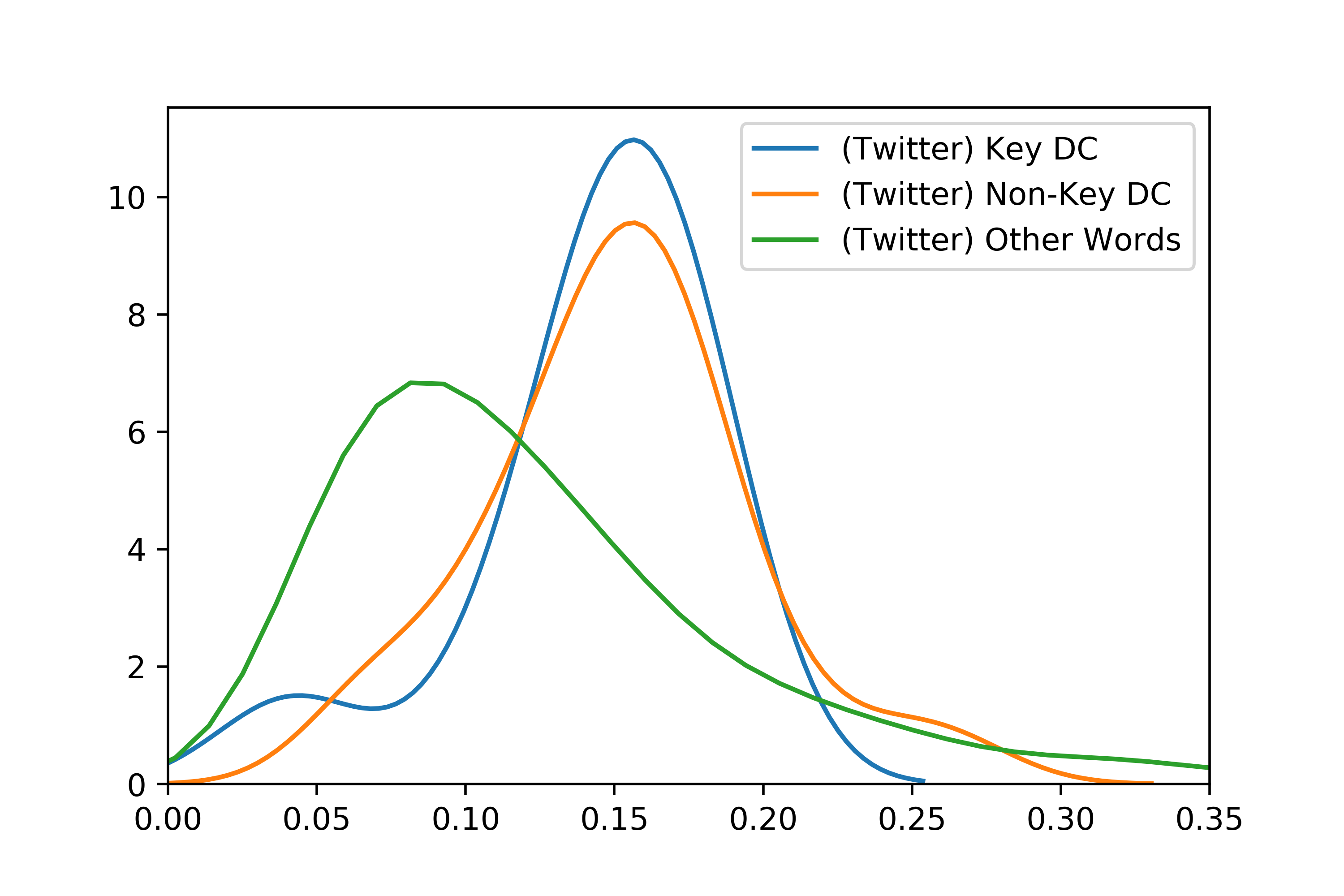} }}
\subfloat[Average Attention Weights on PDTB]{{\includegraphics[width=3in]{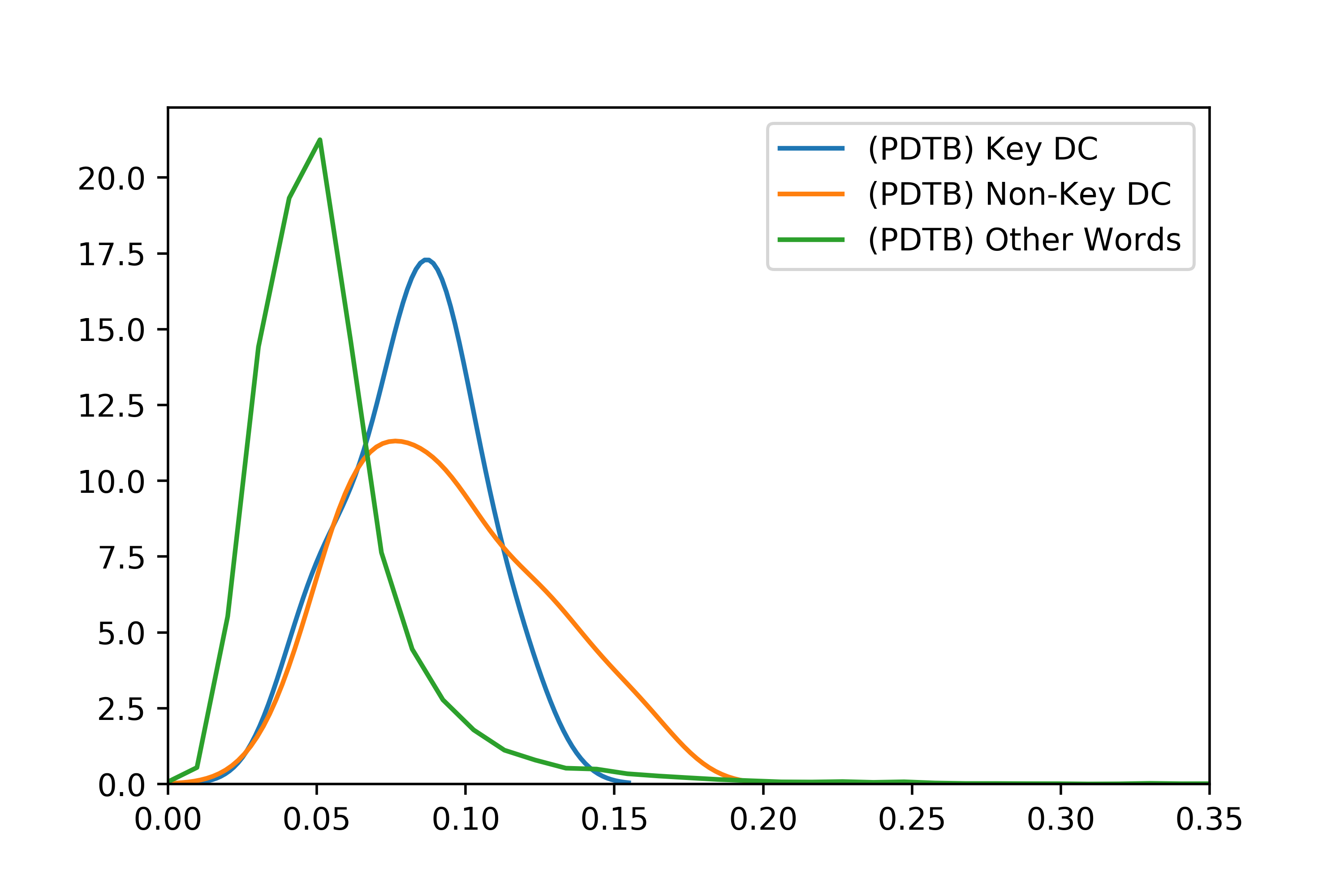} }}

  \caption{Distribution plot with attention weights as a variable in x-axis, `Key DC': discourse connectives used as keywords for the training set collection, `Non-Key DC': discourse connectives which were not included in the keywords. We analyzed the average attention weight distributions of discourse connectives vs other words. Discourse connectives tend to receive higher attention on both PDTB and Twitter\footnotemark.} 
  \label{fig:attn_dist}
\end{figure*}
\footnotetext{Interestingly, on Twitter, the attention weights of social-media-specific variations of `because' obtained similar weights even though the DiscRE model was not systematically designed to capture domain differences of discourse connectives: `because': 0.16, `bcuz': 0.18, `cos': 0.16, `cuz': 0.15, `cause': 0.16.}

\paragraph{Discourse Relation Classification on Social Media.}
To validate DiscRE 
 beyond the existing corpus of newswire domain, it was applied to a discourse relation classification task on our new Twitter discourse relation dataset. We extracted DiscRE, BERT, Ngrams, and Infersent from tweets with the same methods used in the causality task. We conducted 10-fold cross validation and report F1 scores of the models on each class in Table~\ref{tab:Twitter_rel}. The result showed that DiscRE obtains the best performance across all the classes. (Micro F1=0.758).

\subsection{Qualitative Analysis on DiscRE model}

\paragraph{Attention Analysis.}

 First, we ran pretrained DiscRE on the evaluation tweet dataset (Section~\ref{sec:qual_set}) and investigated average attention weights. 
 Discourse connectives gained higher attention than non-discourse-connective words (Figure~\ref{fig:attn_dist}).\footnote{Beyond some outliers due to noisy unigrams and social-media-specific discourse arguments (e.g., emojis or verb phrases with omitted subjects)} 
 This suggests that discourse connectives play a quite significant role in DiscRE. 
 
 Furthermore, we observed that both, the discourse connectives used as keywords for training set collection, as well as the relatively less frequent discourse connectives obtained higher attention weights than other words on the random tweet evaluation set. This pattern supports that our model was not biased towards only prevailing discourse connectives it has already seen from the training set, but generalized quite well on unseen discourse connectives.
 
 When we analyzed attention weights on the DiscRE model for the PDTB dataset, it showed a similar pattern. Although all words in the PDTB vocabulary generally obtained lower attention, the discourse connectives still obtained higher attention weights than other words, and relatively high attention weights were distributed on both keyword and non-keyword discourse connectives in PDTB as well. These results suggest that DiscRE can capture words with important discourse signals even on the other domains.

\paragraph{DiscRE Analysis.}

\begin{figure}[h]
\centering
  \includegraphics[width=3in]{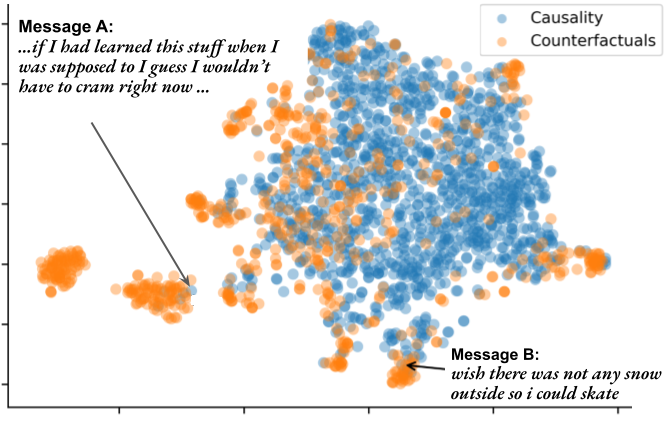}
  \caption{DiscRE differences between counterfactual messages and causality messages. Counterfactual messages are generally positioned at the left side compared to causality messages. When we investigated edge cases of causality messages that clustered closely with counterfactuals, we found causality messages which contained counterfactual relations inside (`Message A': `is doing great.... lol. If I had learned this stuff when I was supposed to I guess I wouldn't have to cram right now. Oh well. There's always next year... or grade 12.' \\ `Message B': `i wish there was not any snow outside so i could skate').} 
  \label{fig:ce_cf_cluster}
\end{figure}

\begin{figure}[h]
\centering
  \includegraphics[width=3in]{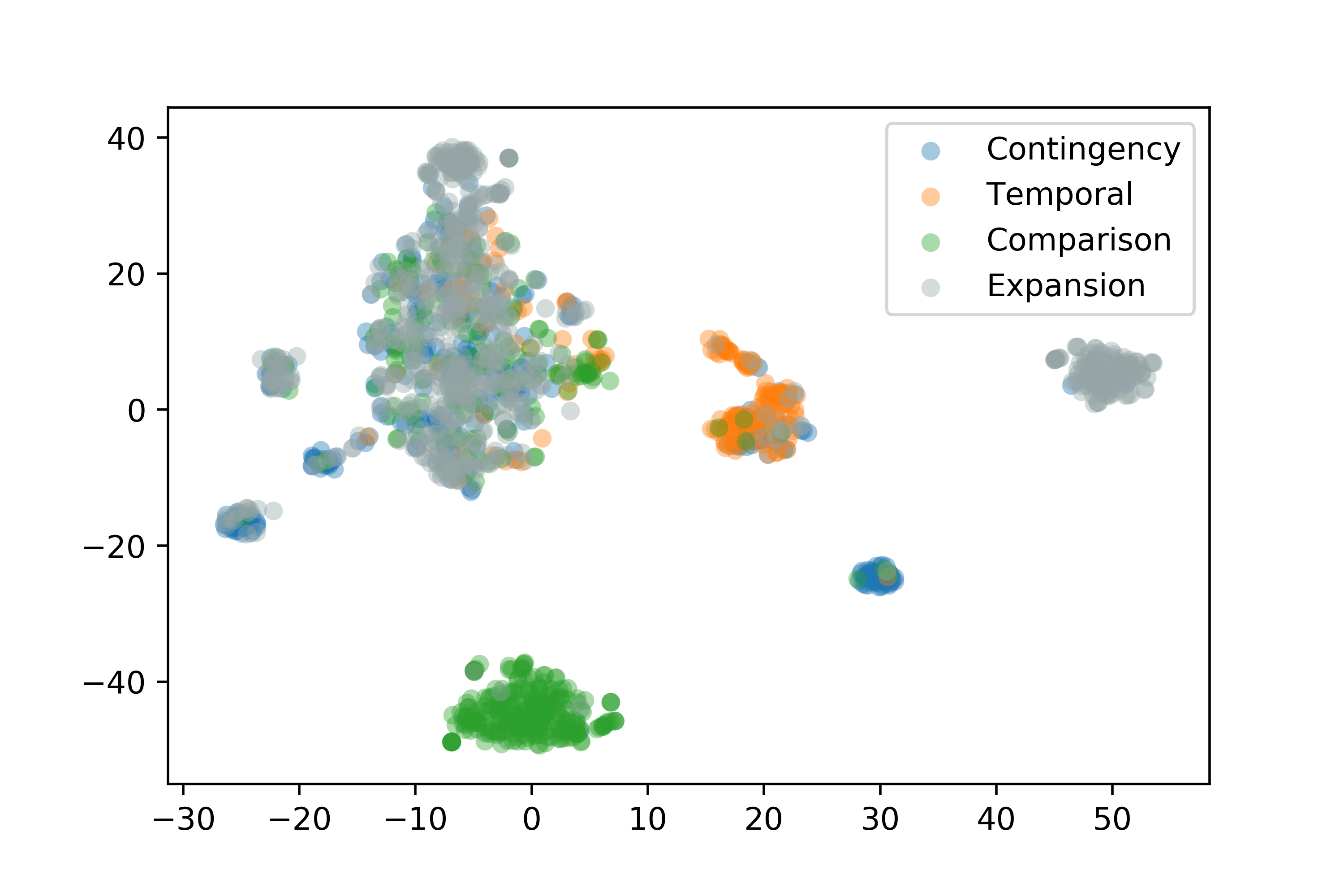}
  \caption{DiscRE differences between the four discourse relation classes of the PDTB dataset. Many examples of implicit discourse relations were clustered on the upper left side. Expansion is a quite general class which may overlap semantically with other types of relations, so they were more widely spread than other relations.} 
  \label{fig:four_way_cluster}
\end{figure}

We evaluated DiscRE on social media discourse relations datasets which are publicly available: causality ~\cite{son2017recognizing} and counterfactual ~\cite{son2018causal}. We averaged the DiscRE embeddings of all adjacent pairs of discourse arguments per message and visualized using tSNE (Figures~\ref{fig:ce_cf_cluster},~\ref{fig:four_way_cluster}). In general, discourse relations are diverse and even the same \textit{Type} show up in various different forms in both explicit and implicit relations, so the distinctions between them are very hard to be captured within just two dimensions. Nevertheless, we found fairly clear patterns that distinguish two different discourse relations; majority counterfactual messages tend to cluster separately towards the left, as compared to causality messages (Figure~\ref{fig:ce_cf_cluster}). \textit{Conjunctive Normal} and \textit{Conjunctive Converse} forms of counterfactuals are especially clustered at the left side separately (e.g., ``I would be healthier, if I had worked out regularly'')~\cite{son2017recognizing}. 

It is noteworthy that the counterfactual relation does not exist as a discourse relation tag in PDTB, but DiscRE still captures its distinguishable properties and even different forms of it (i.e., \textit{Wish verb} forms and \textit{Conjunctive} forms). While this visualization provides significant insights about semantic differences of discourse relations, further analysis over coherent clusters helps us see some discourse-based properties in common (e.g., see `Message A' and `Message B' on Figure~\ref{fig:ce_cf_cluster}).

Additionally, we investigated how well DiscRE can generalize to newswire domain by projecting DiscRE embeddings of discourse relations in the PDTB testset into 2D tSNE, similar to the visualization of causal and counterfatual relations (Figure~\ref{fig:four_way_cluster}). Even though we used most coarse-grained discourse relation classes, DiscRE captured quite coherent patterns of clusters for different relations. Nevertheless, many implicit discourse relations were clustered together on the upper left part as they are generally harder to be captured~\cite{pitler2008easily,rutherford2015improving}.

\section{Conclusion}
This paper suggests a difference in how semantics is modeled in NLP, moving beyond word-level embeddings to embeddings that capture the semantics of discourse relations.
We explored a new task of creating latent discourse relation embeddings, designing a novel weakly supervised multitask learning method and evaluating it both quantitatively and qualitatively over social media and newswire domains. While we built on previous work over discourse relation classes, our results suggest the \textit{continuous} discourse relation embeddings (DiscRE) has certain benefits over manual categorizations. Continuous representations of relations between segments of text have been relatively unexplored yet they can yield subtle attributes of discourse relations, yielding strong performance in applications and perhaps new organizations of functional discourse relations. 

Our model obtained the best performance on the discourse relation classification tasks in both PDTB and our new Twitter discourse dataset. Our model also obtained a new state-of-the-art performance using DiscRE in the social media causal relation prediction task. Further, for predicting discourse relations over PDTB, we found DiscRE achieved the higher performance than other embeddings, suggesting a focus on embedding \textit{relations} 
can capture information not available in other types of modern embeddings which focus on representing particular word or phrase instances rather than their relationships.
We release our dataset, code and pretrained models, for others to explore this new task, better develop continuous representations of discourse relations, as well as to extend discourse relation parsing beyond newswire to other domains.

\section{Limitations}
The model delineated in this work is scalable with large amounts of unsupervised data, but still orders of magnitude less than what modern language models require. The social media validation was performed on a small annotated dataset with a high inter-annotator agreement, limited to 360 tweets that had examples from each relation class .
The model was trained on a single 12GB memory GPU (we used a NVIDIA Titan XP graphics card). 
The approach should be expected to work best with languages that have limited morphology, like English. 

The weakly supervised approach has a small limitation in that it still aligns the model, to some degree, with an existing tagset (i.e. the PDTB discourse relation tagset), but our results suggested we were able to capture relations beyond it (e.g. capturing a relation that is a mix of causal explanation and counterfactuals). 


\section {Ethical Considerations}
All of our work is restricted to document-level information; No user-level information is used. 

\section*{Acknowledgements}
This work was supported by DARPA via Young Faculty Award grant \#W911NF-20-1-0306 to H. Andrew Schwartz at Stony Brook University; the conclusions and opinions expressed are attributable only to the authors and should not be construed as those of DARPA or the U.S. Department of Defense. 
This work was also supported in part by NIH R01 AA028032-01. 


\bibliography{custom}
\bibliographystyle{acl_natbib}

\end{document}